\documentclass[10pt,twocolumn,letterpaper]{article}

\usepackage{times}
\usepackage{epsfig}
\usepackage{graphicx}
\usepackage{amsmath}
\usepackage{amssymb}
\usepackage{booktabs}
\usepackage{subcaption}
\usepackage{algorithm}
\usepackage{algorithmic}

\newcommand{\squeezeup}{\vspace{-2.5mm}}
\begin{document}

\title{Multi-modal Capsule Routing for Actor and Action Video Segmentation Conditioned on Natural Language Queries 
}

\author{
Bruce McIntosh\\
\and
Kevin Duarte\\
\and 
Yogesh S Rawat\\
\and 
Mubarak Shah\\
\and 
{\tt\small \{bwmcint, kevin\_duarte\}@knights.ucf.edu}
\and 
{\tt\small \{yogesh, shah\}@crcv.ucf.edu} 
\and \\
Center for Research in Computer Vision\\
University of Central Florida
}

\maketitle

\begin{abstract}
In this paper, we propose an end-to-end capsule network for pixel level localization of actors and actions present in a video. The localization is performed based on a natural language query through which an actor and action are specified. We propose to encode both the video as well as textual input in the form of capsules, which provide more effective representation in comparison with standard convolution based features. We introduce a novel capsule based attention mechanism for fusion of video and text capsules for text selected video segmentation. The attention mechanism is performed via joint EM routing over video and text capsules for text selected actor and action localization. The existing works on actor-action localization are mainly focused on localization in a {\bf \textit{single}} frame instead of the full video. Different from existing works, we propose to perform the localization on {\bf \textit{all}} frames  of the video. To validate the potential of the proposed network for actor and action localization on all the frames of a video, we extend an existing actor-action dataset (A2D) with annotations for all the frames. The experimental evaluation demonstrates the effectiveness of the proposed capsule network for text selective actor and action localization in videos, and it also improves upon the performance of the existing state-of-the art works on single frame-based localization.
   
\end{abstract}

\begin{figure}[t]
\begin{center}
\includegraphics[width=1\linewidth]{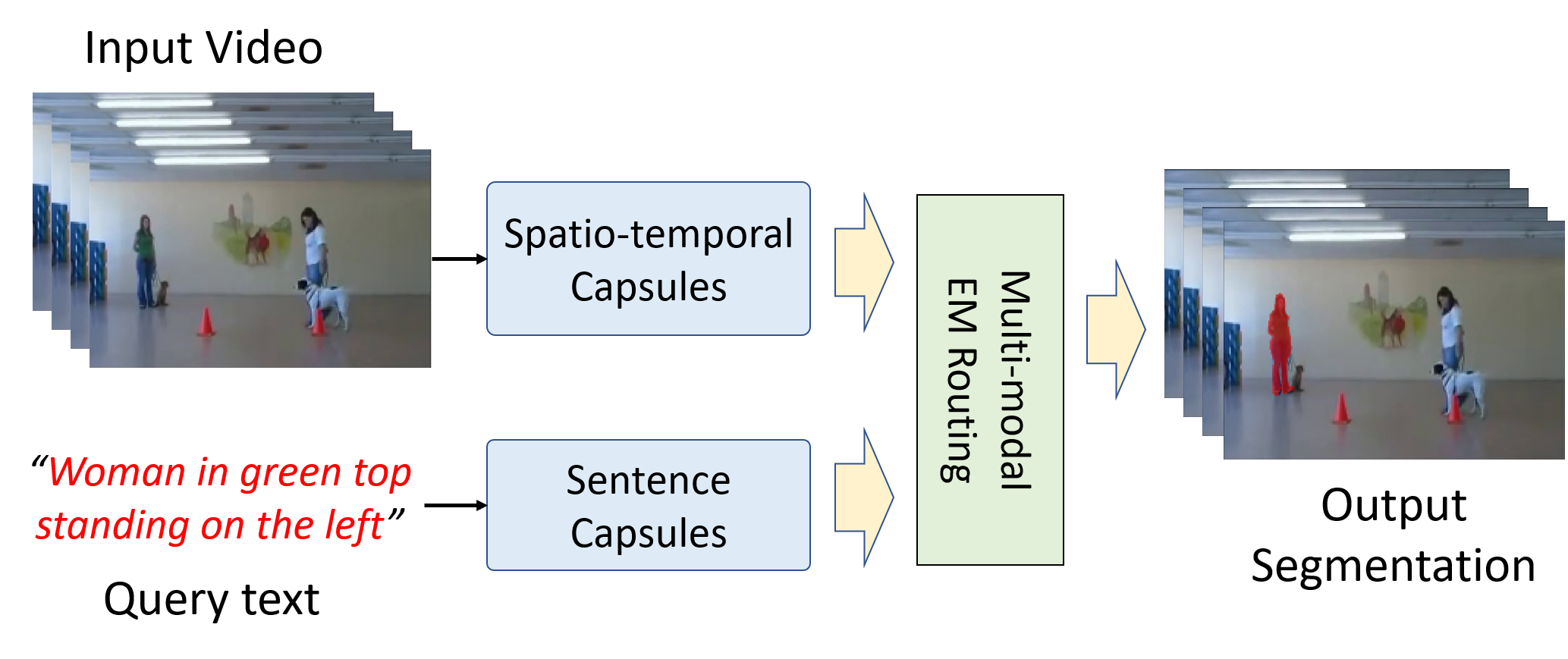}
\end{center}
   \caption{Overview of the proposed approach. For a given video, we want to localize the actor and action which are described by an input textual query. Capsules are extracted from both the video and the textual query, and a joint EM routing algorithm creates high level capsules, which are further used for localization of selected actors and actions. }
\label{fig:overview}
\end{figure} 
\section{Introduction and Related Work}


Video semantic segmentation is an active topic of research in computer vision,  as it serves as a basic foundation for various real-world applications, such as autonomous driving. It is a very challenging problem where every pixel in the video needs to be assigned a semantic category. Video action detection is a subset of this problem, where we are only interested in segmenting/localizing the actors and objects involved in the activities present in a video. The problem becomes even more interesting, when we want to focus on only selective actors in the video. In this work, we focus on selective localization of actors and actions in a video based on a textual query, as seen in Figure \ref{fig:overview}.

We have recently seen good progress in the task of action localization \cite{arsurvey, tcnn, ava, tubelets, twostream, carreira2017quo, duarte2018videocapsulenet}, and it is mainly accredited to the success in deep learning along with the availability of large scale datasets \cite{ava, ucf101, jhmdb}. However, one limitation of these datasets is that the actor is mainly a person performing various activities. Xu et al. \cite{xu2015can} introduce an actor-action video dataset (A2D), which has several actor and action pairs; this dataset presents several challenges as there could be different types of actors besides just humans performing the actions. Also, there could be {\em multiple} actors present in a scene, which is quite challenging when compared to typical datasets, where we have only  {\em one} actor performing the action. 
Their experiments on A2D showed that a joint inference over actor and action outperforms methods that treat them independently. Ji et al. \cite{ji2018end} explored the role of optical flow along with RGB data and proposed an end-to-end deep network for joint actor and action localization.

Gavrilyuk et al. \cite{gavrilyuk2018actor} recently extended the A2D dataset with human generated sentences, describing the actors and actions in the video, and proposed the task of actor and action segmentation from a sentence. Their method uses a convolutional network for both visual as well as textual inputs and predicts localization on {\em one frame} of the video. We propose a different approach, where we make use of {\em capsules} for both visual as well as text encoding, and perform localization on the {\em full video} instead of just one frame, in order to fully utilize the spatiotemporal information captured by the video.


Hinton et al. first introduced the idea of capsules in \cite{hinton2011transforming}, and subsequently capsules were popularized in \cite{sabour2017dynamicrouting}, where dynamic routing for capsules was proposed. This was further extended in \cite{hinton2018emrouting}, where a more effective EM routing algorithm was introduced. Recently, capsule networks have shown state-of-the-art results for human action localization in video \cite{duarte2018videocapsulenet}, object segmentation in medical images \cite{lalonde2018capsules}, and text classification \cite{zhao2018investigating}. We propose to extend the use of capsule networks into the multi-modal domain, where the segmentation and localization of objects in video are conditioned on a natural language input. We introduce a novel capsule based attention mechanism for fusion of video and text capsules for text selected segmentation. 

There are several existing works focusing on learning the interaction between text and visual data \cite{li2017person, yamaguchi2017spatio, gao2017tall, hendricks2017localizing}. These works are mainly focused on whole image and video level detections as opposed to pixel level segmentation. Hu et al. \cite{hu2016segmentation} introduced the problem of segmenting images based on a natural language expression; their method for merging images and text in a convolutional neural network (CNN) was by concatenating features extracted from both modalities and followed  by a convolution. Li et al. \cite{li2017tracking} propose a different approach to merge these two modalities for the task of tracking a target in a video; they use an element-wise multiplication between the image feature maps and the sentence features in a process called dynamic filtering. Our proposed method encodes both the video as well as the textual query as capsules, and makes use of an EM routing algorithm to learn the interaction between text and visual information. The method differs from conventional convolutional approaches as it takes advantage of capsules networks' ability to model entities and their relationships; the routing procedure allows our network to learn the relationship between entities from different modalities. 


\begin{figure*}[t]
\centering
\includegraphics[width=0.9\linewidth]{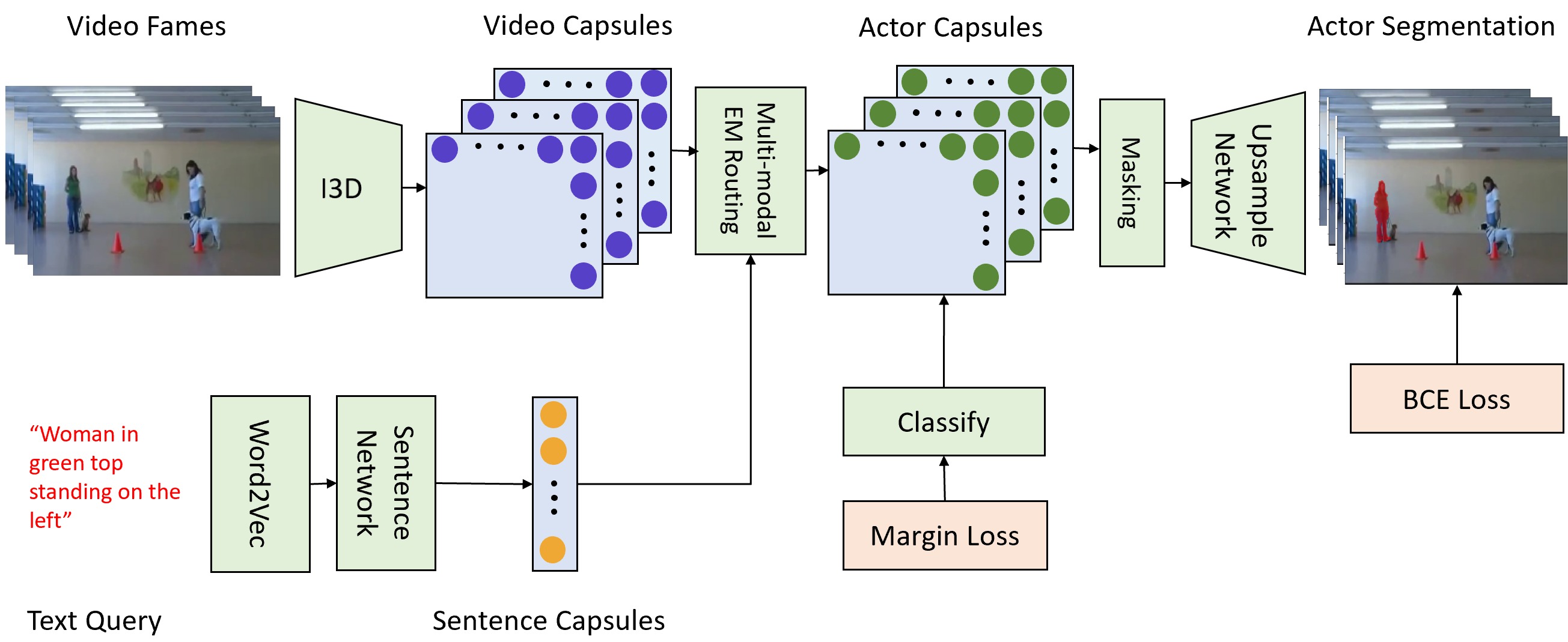}
   \caption{Network Architecture. Capsules containing spatiotemporal features are created from video frames, and capsules representing a textual query are created from natural language sentences. These capsules are routed together to create capsules representing actors in the image. The actor capsule poses go through a masking procedure and an upsampling network to create binary segmentation masks of the actor specified in the query.}
\label{fig:network}
\end{figure*} 

In summary, we make the following contributions in this work: 
    (1) We propose an end-to-end capsule network for the task of selective actor and action localization in videos, which encodes both the video and the textual query in the form of capsules.
    (2) We introduce a novel multi-modal conditioning routing as an attention mechanism to address the issue of cross-modality capsule selection.
    (3) To demonstrate the potential of the proposed text selective actor and action localization in videos we extend the annotaions in A2D dataset to full video clips. Our experiments demonstrate the effectiveness of the proposed method, and we show its advantage over the existing state-of-the art works in terms of performance.

\section{Conditioning with Multi-modal Capsule Routing} \label{sect-multimodal}

We argue that capsule networks can effectively perform multi-modal conditioning. Capsules represent entities and routing uses \textit{high-dimensional coincidence filtering} \cite{hinton2018emrouting} to learn part-to-whole relationships between these entities. There are several possible ways to incorporate conditioning into capsule networks. One trivial approach would be to apply a convolutional method (concatenation followed by a 1x1 convolution \cite{hu2016segmentation} or multiplication/dynamic filtering \cite{li2017tracking}) to create conditioned feature maps, and then extract a set of capsules from these feature maps. This, however, would not perform much better than the fully convolutional networks, since the same conditioned feature maps are obtained from the merging of the visual and textual modalities, and the only difference is how they are transformed into segmentation maps.

Another method would be to first extract a set of capsules from the video, and then apply the dynamic filtering on these capsules. This can be done by (1) applying a dynamic filter to the pose matrices of the capsules, or (2) applying a dynamic filter to the activations of the capsules. The first is not much different than the trivial approach described above, since the same set of conditioned features would be present in the capsule pose matrices, as opposed to the layer prior to the capsules. The second approach would just discount importance of the votes corresponding to entities not present in the sentence; this is not ideal, since it does not take advantage of routing's ability to find agreement between entities in both modalities.  


Instead, we propose an approach that leverages the fact that the same entities exist in both the video and sentence inputs and that routing can find similarities between these entities. From the video, we extract a grid of capsules describing the visual entities, $C_v$, with pose matrices $M_v$ and activations $a_v$. Similarly, from the sentence, we generate sentence capsules, $C_s$, with pose matrices $M_s$ and activations $a_s$. Each set of capsules has transformation matrices $T_{vj}$ and $T_{sj}$, for video and text respectively, which are used to cast votes for the capsules in the following layer. Video capsules at different spatial locations share the same transformation matrices. Using the procedure described in Algorithm \ref{alg:merging}, we obtain a grid of higher-level capsules, $C_j$. This algorithm allows the network to find similarity, or agreement, between the votes of the video and sentence capsules at every location on the grid. If there is agreement between the votes, then the same entity exists in both the sentence and the given location in the video, leading to a high activation of the capsule corresponding to that entity. Conversely, if the sentence does not describe the entity present at the given spatial location, then the activation of the higher-level capsules will be low since the votes would disagree.


\begin{algorithm}
\caption{{\em Multi-modal Capsule Routing}. The $\{\text{\textbullet} ; \text{\textbullet}\}$ operation is concatenation, such that the activations and votes of both the video and sentence capsules are inputs to the EM routing procedure described in \cite{hinton2018emrouting}. }
\label{alg:merging}
\begin{algorithmic}
    \STATE $V_{sj} \leftarrow M_k T_{sj}$
    \FOR{$x=1 \textbf{ to } W$}
        \FOR{$y=1 \textbf{ to } H$}
            \STATE $V_{vj} \leftarrow M_v \left[ x, y \right] T_{vj} $
            \STATE $C_j \left[ x, y \right] \leftarrow \text{EM-Routing} \left( \{ a_s; a_v\left[ x, y \right] \}, \{ V_{sj}; V_{vj} \}\right)$
        \ENDFOR
    \ENDFOR
    \RETURN $C_j$
\end{algorithmic}
\end{algorithm}

This formulation of multi-modal conditioning using capsules allows the network to learn a set of entities (capsules) from both the visual and sentence inputs. Then, the voting and routing procedure allows us to find agreement between both modalities in the form of higher-level capsules. Suppose there is a higher-level capsule describing a \textit{dog}. When given a query like ``The brown dog running" the sentence capsules' vote matrices corresponding to the dog class, contain the different properties of the dog, like the fact that it is running or that it is brown. If there is a \textit{running brown dog} at some location in the visual input, then the visual votes at that location would be similar to the the sentence's votes,  so the activation of the higher-level dog capsule would be high at that location. If, for instance, there is a \textit{black dog} or a \textit{dog rolling} at some location then the votes would not agree, and activation for the dog capsule would be low there.

Although this work mainly focuses on video and text, the multi-modal capsule routing procedure can be applied to capsules generated from many other modalities, like images or audio. 

\section{Network Architecture}
The overall network architecture is shown in Figure \ref{fig:network}. In this section, we discuss the various components of the architecture as well as the objective function used to train the network.

\subsection{Video Capsules} 
\label{l_vid_cap}
The video input consists of 4 $224 \times 224$ frames. The process for generating video capsules begins with an Inception based 3D convolutional network known as I3D \cite{carreira2017quo}, which generates 832 - $28 \times 28$ spatiotemporal feature maps taken from the maxpool3d\_3a\_3x3 layer. Capsule pose matrices and activations are generated by applying a $9 \times 9$ convolution operation to these feature maps, with linear and sigmoid activations respectively. Since there is no padding for this operation, the result is a $20 \times 20$ capsule layer with 8 capsule types.

\begin{figure*}[t]
\centering
\includegraphics[width=0.9\linewidth]{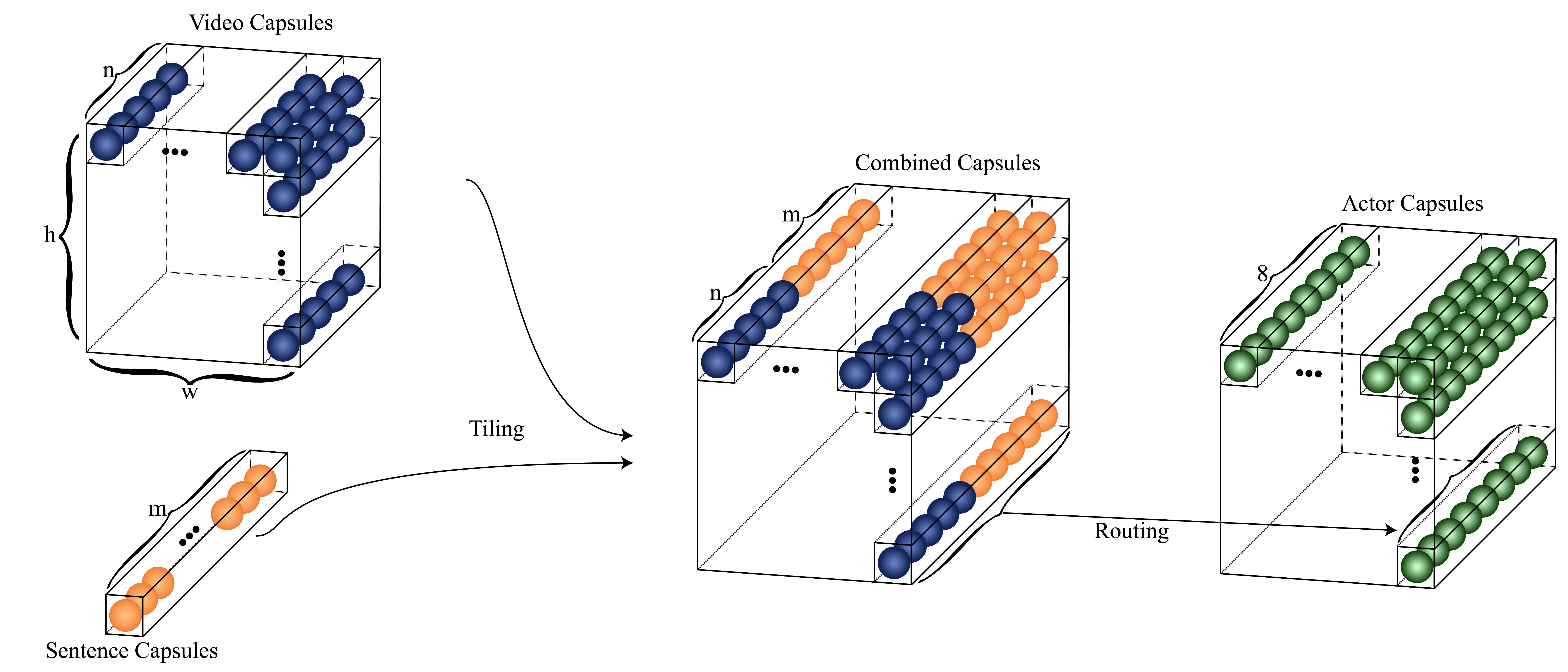}
   \caption{Capsule Merging. Video capsules are created with n types of capsules at each pixel location. Sentence capsules are created with m types of capsules for the sentence. The sentence capsules are tiled over all of the spatial locations. EM routing is performed to create the next higher level of capsules representing actors at each spatial location. } 
\label{fig:merging}
\end{figure*} 

\subsection{Sentence Capsules} 
A series of convolutional and fully connected layers is used to generate the sentence capsules. First, each word from the sentence is converted into a size 300 vector using a word2vec model pre-trained on the Google News Corpus \cite{mikolov2013w2v}. Sentences in the network are set to be 16 words, so longer sentences are truncated, and shorter sentences are padded with zeros. The sentence representation is then passed through 3 parallel stages of 1D convolution with kernel sizes of 2, 3 and 4 with a ReLU activation. We then apply max-pooling to obtain 3 vectors, which are concatenated and passed through a max-pooling layer to obtain a single length 300 vector to describe the entire sentence. A fully connected layer then generates the 8 pose matrices and 8 activations for the capsules which represent the entire sentence. We found that this method of generating sentence capsules performed best in our network: various other methods are explored in the Supplementary Material.

\subsection{Merging and Masking}
Once the video and sentence capsules are obtained, we merge them in the manner described in section \ref{sect-multimodal}, and depicted in Figure \ref{fig:merging}. The result of the routing operation is a $20 \times 20$ grid with 8 capsule types - one for each actor class in the A2D dataset and one for a ``background" class, which is used to route unnecessary information. The activations of these capsules correspond to the existence of the corresponding actor at the given location, so averaging the activations over all locations gives us a classification prediction over the video clip. We perform the capsule masking procedure described in \cite{sabour2017dynamicrouting}. When training the network, we mask (multiply by 0) all pose matrices not corresponding to the ground truth class. At test time, we mask the pose matrices not corresponding to the predicted class. These masked poses are then fed into an upsampling network to generate a foreground/background segmentation mask for the actor described by the sentence.

\subsection{Upsampling Network}
The upsampling network consists of 5 convolutional transpose layers. The first of these increases the feature map dimension from $20 \times 20$ to $28 \times 28$ with a $9 \times 9$ kernel, which corresponds to the $9 \times 9$ kernel used to create the video capsules from the I3D feature maps. The following 3 layers have $3 \times 3 \times 3$ kernels and are strided in both time and space, so that the output dimensions are equal to the input video dimensions ($4 \times 224 \times 224$). The final segmentation is produced by a final layer which has a $3 \times 3 \times 3$ kernel. We must note here, that this method diverges from previous methods in that it outputs segmentations for all input frames, rather than a single frame segmentation per video clip input. We use parameterized skip connections from the I3D encoder to obtain more fine-grained segmentations. At each step of upsampling, lower resolution segmentation maps are generated to aid in the training of these skip connections. 

\subsection{Objective Function} 
The network is trained end-to-end using an objective function based on classification and segmentation losses. For classification, we use a spread loss which is computed as follows:
\begin{equation}
    L_c = \sum_{i \neq t} {\max {\left(0, m - \left( a_{gt} - a_i \right) \right)}^2},
\end{equation}
where $m \in \left( 0, 1 \right)$ is a margin, $a_i$ is the activation of the capsule corresponding to class $i$, and $a_{gt}$ is the activation of the capsule corresponding to the ground-truth class. During training, $m$ is linearly increased between $0.2$ and $0.9$.

The segmentation loss is computed using sigmoid cross entropy. When averaged over all $N$ pixels in the segmentation map, we get the following loss: 
\begin{equation}
    L_s = -\frac{1}{N}\sum_{j=1}^N p_j \log \left( \hat{p}_j \right) - \left( 1 - p_j \right) \log \left( 1 - \hat{p}_j \right),
\end{equation}
where $p_j \in \{0, 1 \}$ is the ground-truth segmentation map and $\hat{p}_j \in \left[ 0, 1\right]$ is the network's output segmentation map. We use this segmentation loss at several resolutions to aid in the training of the skip connections.

The final loss is a weighted sum between the classification and segmentation losses:
\begin{equation}
    L = \lambda L_c + \left( 1 - \lambda \right) L_s,
\end{equation}
where $\lambda$ is set to $0.5$ when training begins. Since the network quickly learns to classify the actor when given a sentence input, we set $\lambda$ to $0$ when the classification accuracy saturates (over 95\% on the validation set). We find that this reduces over-fitting and results in better segmentations.

\section{Experiments}

\paragraph{Implementation Details} The network was implemented using PyTorch \cite{paszke2017automatic}. The I3D used weights pretrained on Kinetics \cite{kay2017kinetics} and fine tuned on Charades \cite{sigurdsson2016hollywood}. The network was trained using the adam optimizer \cite{kingma2014adam} with a learning rate of .001. As video resolutions vary within different datasets, all video inputs are scaled to $224 \times 224$ while maintaining aspect ratio through the use of horizontal black bars. When using bounding box annotations, we consider pixels within the bounding box to be foreground and pixels outside of the bounding box to be background.

\begin{table*}
\small
\centering
\begin{tabular}{l ccccc c cc}
\toprule
 & \multicolumn{5}{c}{\bf Overlap} & \multicolumn{1}{c}{\bf mAP} & \multicolumn{2}{c}{\bf IoU} \\
 & P@0.5 & P@0.6 & P@0.7 & P@0.8 & P@0.9 & 0.5:0.95 & Overall & Mean \\
\cmidrule(lr){2-6} \cmidrule(lr){7-7} \cmidrule(lr){8-9} 
Hu et al. \cite{hu2016segmentation}           & 34.8  & 23.6  & 13.3  & 3.3   & 0.1   & 13.2  & 47.4  & 35.0 \\
Li et al.  \cite{li2017tracking}         & 38.7  & 29.0  & 17.5  & 6.6   & 0.1   & 16.3  & 51.5  & 35.4 \\
Gavrilyuk et al. \cite{gavrilyuk2018actor}   & 50.0  & 37.6  & 23.1  & 9.4   & 0.4   & 21.5  & 55.1  & 42.6 \\
\midrule
Our Network         &\bf 52.6  &\bf 45.0  &\bf 34.5  &\bf 20.7  &\bf 3.6  &\bf 30.3  &\bf 56.8  &\bf 46.0 \\ 

\bottomrule
\end{tabular}
\caption{Results on A2D dataset with sentences. Baselines \cite{hu2016segmentation,li2017tracking} take only single image/frame inputs. Gavrilyuk et al. \cite{gavrilyuk2018actor} uses multi-frame RGB and Flow inputs. Our model uses only multi-frame RGB inputs and outperforms other state-of-art-methods in all metrics without the use of optical flow. }
\label{a2dsentences-table}
\end{table*}

\begin{table*}
\small
\centering
\begin{tabular}{l ccccc c cc}
\toprule
 & \multicolumn{5}{c}{\bf Overlap} & \multicolumn{1}{c}{\bf mAP} & \multicolumn{2}{c}{\bf IoU} \\
 & P@0.5 & P@0.6 & P@0.7 & P@0.8 & P@0.9 & 0.5:0.95 & Overall & Mean \\
\cmidrule(lr){2-6} \cmidrule(lr){7-7} \cmidrule(lr){8-9} 
Hu et al. \cite{hu2016segmentation} & 63.3 & 35.0 & 8.5 & 0.2 & 0.0 & 17.8 & 54.6 & 52.8  \\
Li et al. \cite{li2017tracking} & 57.8 & 33.5 & 10.3 & 0.6 & 0.0 & 17.3 & 52.9 & 49.1 \\
Gavrilyuk et al. \cite{gavrilyuk2018actor} & \bf 69.9 & 46.0 & 17.3 & 1.4 & 0.0 & 23.3 & \bf 54.1 & \bf 54.2 \\
\midrule
Our Network         & 63.8  &\bf 47.9  & \bf26.3  & \bf4.0  & 0.0  & \bf24.3 & 49.2  & 52.0 \\   
\bottomrule
\end{tabular}
\caption{Results on JHMDB dataset with sentences. Our model outperforms other state-of-the-art methods at higher IoU thresholds and in the mean average precision metric.}
\label{jhmdb-table}
\end{table*}



\subsection{Single-Frame Segmentation Conditioned on Sentences}
In this experiment, a video clip and a human generated sentence describing one of the actors in the video are taken as inputs, and the network generates a binary segmentation mask localizing the described actor. Similar to previous methods, the network is trained and tested on the single frame annotations provided in the A2D dataset. To compare our method with previous approaches, we modify our network in these experiments. We replace the 3d convolutional transpose layers in our upsampling network to 2d convolutional transpose layers to output a single frame segmentation.
\squeezeup

\begin{figure}[t!]
\begin{center}

\includegraphics[width=1.0\linewidth]{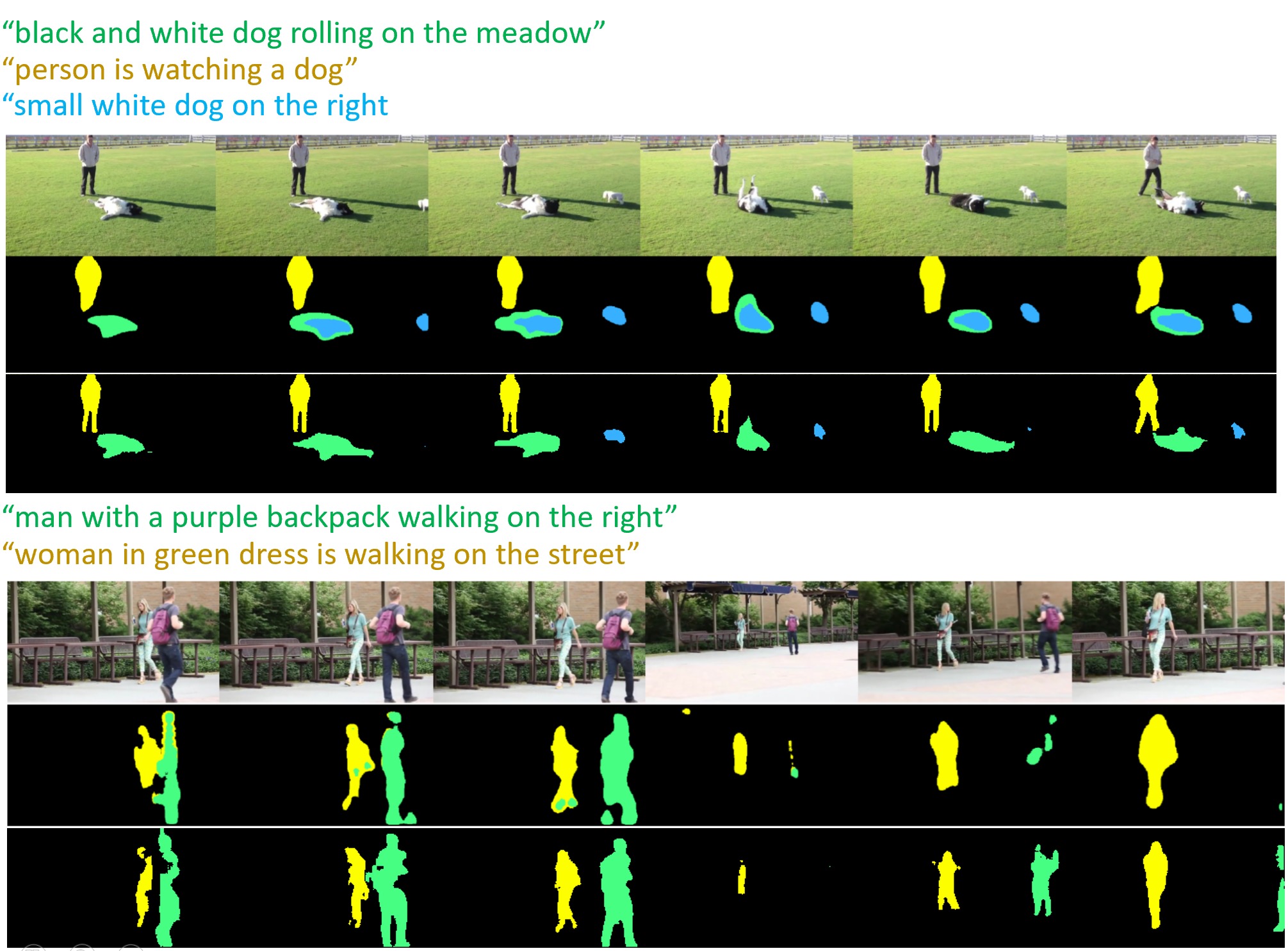}
\end{center}
   \caption{A comparison of our results with \cite{gavrilyuk2018actor}. The sentence query colors correspond with the segmentation colors. The first row are frames from the input video. The second row shows the segmentation output from \cite{gavrilyuk2018actor}, and the third row shows the segmentation output from our model. In both examples, our model produces more finely detailed output, where the separation of the legs can be clearly seen. Our model also produces an output that is more accurately conditioned on the sentence query, as seen in the first example where our network segments the correct dog for each query, while \cite{gavrilyuk2018actor} incorrectly selects the center dog for both queries.   
   }
\label{fig:qual-comparison}
\end{figure}

\paragraph{Datasets} We conduct our experiments on two datasets: A2D \cite{xu2015can} and J-HMDB \cite{jhmdb}.  The A2D dataset contains 3782 videos (3036 for training and 746 for testing) consisting of 7 actor classes, 8 action classes, and an extra action label \textit{none}, which accounts for actors in the background or actions different from the 8 action classes. Since actors cannot perform all labeled actions, there are a total of 43 valid actor-action pairs. Each video in A2D has 3 to 5 frames which are annotated with pixel-level actor-action segmentations. The J-HMDB dataset contains 928 short videos with 21 different action classes. All frames in the J-HMDB dataset are annotated with pixel-level segmentation masks. Gavrilyuk et al. \cite{gavrilyuk2018actor} extended both of these datasets with human generated sentences that describe the actors of interest for each video. These sentences use the actor and action as part of the description, but many do not include the action and rely on other descriptors such as location or color.
\squeezeup
\paragraph{Evaluation} We evaluate our results using all metrics used in \cite{gavrilyuk2018actor}. The \textit{overall IoU} is the intersection-over-union (IoU) over all samples, which tends to favor larger actors and objects. The \textit{mean IoU} is the IoU averaged over all samples, which treats samples of different sizes equally. We also measure the precision at 5 IoU thresholds and the mean average precision over $.50 : .05 : .95$ \cite{lin2014microsoft}.
\squeezeup
\paragraph{Results} We compare our results on A2D with previous approaches in Table \ref{a2dsentences-table}. Our network outperforms previous state-of-the-art methods in all metrics, and has a notable 9\% improvement in the mAP metric, even though we do not process optical flow, which would require extra computation. We also find that our network achieves much stronger results at higher IoU thresholds, which signifies that the segmentations produced by the network are more fine-grained and adhere to the contours of the queried objects. Qualitative results on A2D can be found in Figure \ref{fig:qual-comparison}.

Following the testing procedure in \cite{gavrilyuk2018actor}, we test on all the videos of J-HMDB using our model trained on A2D without fine-tuning. The results on J-HMDB are found in Table \ref{jhmdb-table}; our network outperforms other methods at the higher IoU thresholds (0.6, 0.7, and 0.8) and in the mAP metric. 

\subsection{Full Video Segmentation Conditioned on Sentences}
In this set of experiments, we train the network using the bounding box annotations for all the frames. Since previous baselines only output single frame segmentations, we test our method against our single-frame segmentation network as a baseline. It can generate segmentations for an entire video, by processing the video frame-by-frame.

\begin{figure}[t!]
\centering
\includegraphics[width=1.0\linewidth]{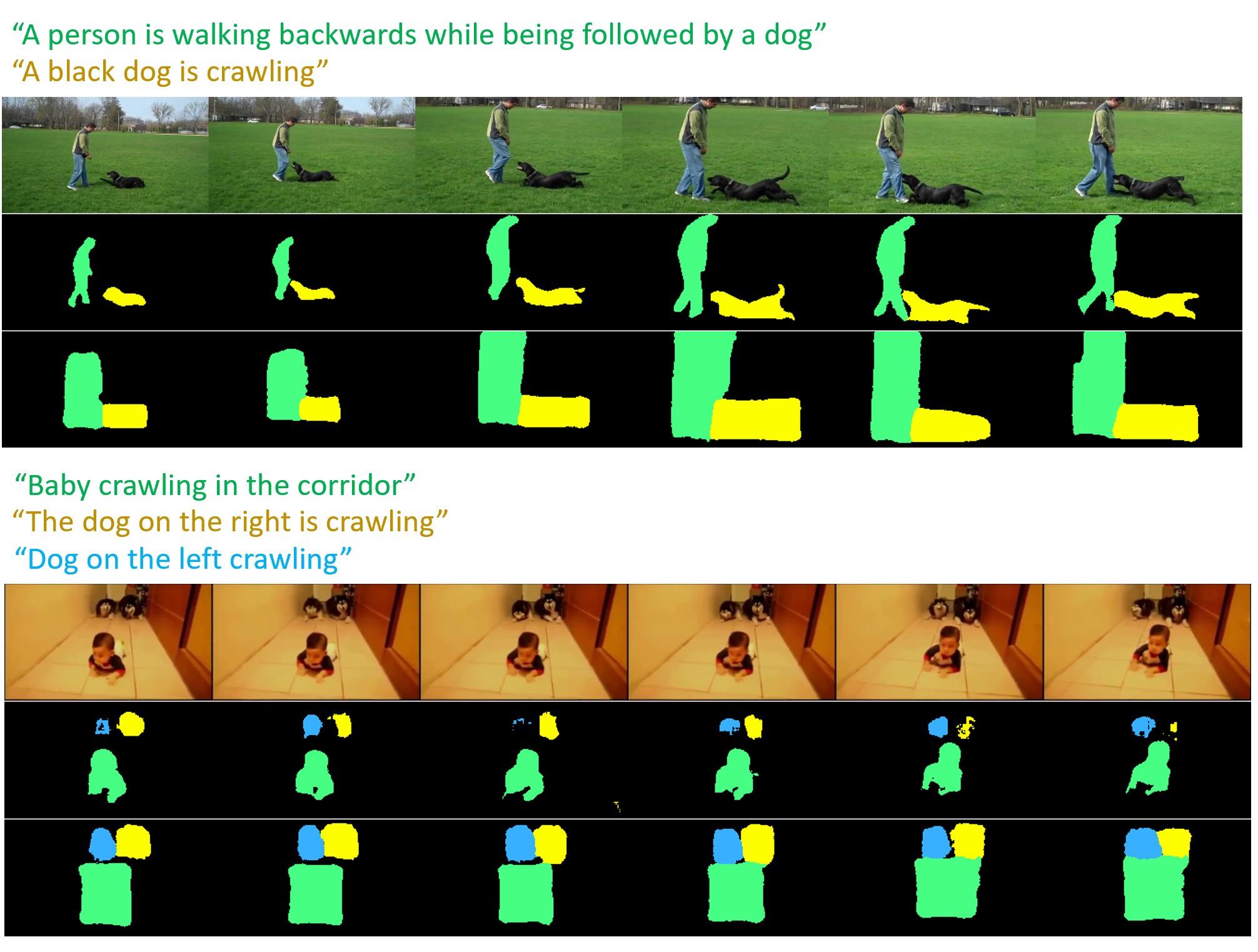}
   \caption{Qualitative results. The sentence query colors correspond with the segmentation colors. The first row are frames from the input video. The second row contains the segmentations from the network trained only using pixel-wise annotations, and the third row contains the segmentations from the network trained using bounding box annotations on all frames. The segmentations from the network trained using bounding boxes are more box-like, but the extra training data leads to fewer missegmentations or under-segmentations as seen in the second example. Higher resolution qualitative results, with their corresponding ground-truths can be found in the Supplementary Material.} 
\label{fig:qual-bbox}
\end{figure}

\squeezeup
\paragraph{Datasets} We extend the  A2D dataset by adding bounding box localizations for the actors of interest in every frame of the dataset. This allows us to train and test our method using the entire video and not just 3 to 5 frames per video. The J-HMDB dataset has annotations on all frames, so we can evaluate the method on this dataset as well.
\squeezeup
\paragraph{Evaluation} To evaluate the segmentation results for entire videos, we consider each video as a single sample. Thus, the IoU computed is the intersection-over-union between the ground-truth tube and the generated segmentation tube.  Using this metric, we can calculate the \textit{video overall IoU} and the \textit{video mean IoU}; the former will favor both larger objects and objects in longer videos, while the latter will treat all videos equally. We also measure the precision at 5 different IoU thresholds and the video mean average precision over $.50 : .05 : .95$.
\squeezeup
\paragraph{Results} Since the network is trained using the bounding box annotations, the produced segmentations are more block-like, but it is still able to successfully segment the actors described in given queries. We compare the qualitative results between the network trained only using fine-grained segmentations and the network trained using bounding box annotations in Figure \ref{fig:qual-bbox}. When tested on the A2D dataset, we find that there is a significant improvement in all metrics when compared to the network trained only on single frames with pixel-wise segmentations. However, this is to be expected, since the ground-truth tubes are bounding boxes and box-like segmentations around the actor would produce higher IoU scores. For a fairer comparison, we place a bounding box around the fine-grained segmentations produced by the network trained on the pixel-wise annotations: this produces better results since the new outputs more resemble ground-truth tubes. Even with this change, the network trained on bounding box annotations has the strongest results since it learned all frames in the training videos, as opposed to a handful of frames per video. The results of this experiment can be seen in Table \ref{bbox-table}.

The J-HMDB dataset has pixel-level annotations for all frames, so the box-like segmentations produced by the network should be detrimental to results; we found that this was the case: the network performed poorly in every metric when compared to the network trained on fine-grained pixel-level annotations..

\subsection{Image Segmentation Conditioned on Sentences}
We also evaluate our method by segmenting images based on text queries. To make as few modifications to the network as possible, the single images are repeated to create a ``boring" video input with 4 identical frames.
\squeezeup
\paragraph{Dataset} We use train and test on the ReferItGame dataset \cite{kazemzadeh2014referitgame}, which contains 20000 images with 130525 natural language expressions describing various objects in the images. We use the same train/test splits as \cite{hu2016segmentation, Shi_2018_ECCV}, with 9000 training images and 10000 testing images. Unlike A2D there are no predefined set of actors, so no classification loss or masking is used.
\squeezeup
\squeezeup
\paragraph{Results} The results for this experiment can be seen in Table \ref{referit-table}. We obtain similar results to other state-of-the-art approaches, even though our network architecture is designed for actor/action video segmentation. This demonstrates that our proposed capsule routing procedure is effective on multiple visual modalities - both videos and images.

\begin{table*}[h]
\small
\centering
\begin{tabular}{l ccccc c cc} 
\toprule
 & \multicolumn{5}{c}{\bf Video Overlap} & \multicolumn{1}{c}{\bf v-mAP} & \multicolumn{2}{c}{\bf Video IoU} \\
 & P@0.5 & P@0.6 & P@0.7 & P@0.8 & P@0.9 & 0.5:0.95 & Overall & Mean \\
\cmidrule(lr){2-6} \cmidrule(lr){7-7} \cmidrule(lr){8-9} 
Key frames (pixel)       & 9.6  & 1.6  & 0.4  & 0.0  & 0.0  & 1.8  & 34.4  & 26.6 \\
Key frames (bbox)        & 41.9  & 33.3  & 22.2  & 10.0  & 0.1  & 21.2  & 51.5  & 41.3 \\
All frames         & 45.6  & 37.4  & 25.3  & 10.0  & 0.4  & 23.3  & 55.7  & 41.8 \\  
\bottomrule
\end{tabular}
\caption{Results on A2D dataset with bounding box annotations. The first row is for the network trained with only pixel-level annotations on key frames of the video, and evaluated with its pixel-wise segmentation output. The second is the same network, but a bounding-box is placed around its segmentation output for evaluation. The final row, is the network trained with bounding box annotations on all frames. Significant performance gain is achieved when training with all frames.} 
\label{bbox-table}
\end{table*}

\begin{table}
\small
\centering
\begin{tabular}{l cccc}
\toprule
 & Overall IoU & P@0.5 & P@0.7 & P@0.9   \\
\cmidrule(lr){2-5}
Hu et al. \cite{hu2016segmentation} & 56.83 & 43.86 & 26.65 & 6.47  \\
Shi et al. \cite{Shi_2018_ECCV} & 59.09 & 45.87 & 32.82 & 11.79  \\
\midrule
Our Network         & 55.7  & 43.4   & 28.3  & 9.7   \\   
\bottomrule
\end{tabular}
\caption{Results on ReferItGame dataset. This result for \cite{hu2016segmentation} is obtained by using Deeplab101 as a backbone network, as described in \cite{Shi_2018_ECCV}. We achieve comparable results, even with a network designed for video inputs. This level of performance can be partially attributed to the lack of classification loss and masking, which tends to improve segmentation results in our network.}
\label{referit-table}
\end{table}

\begin{figure}[t]
\centering
\includegraphics[width=1.0\linewidth]{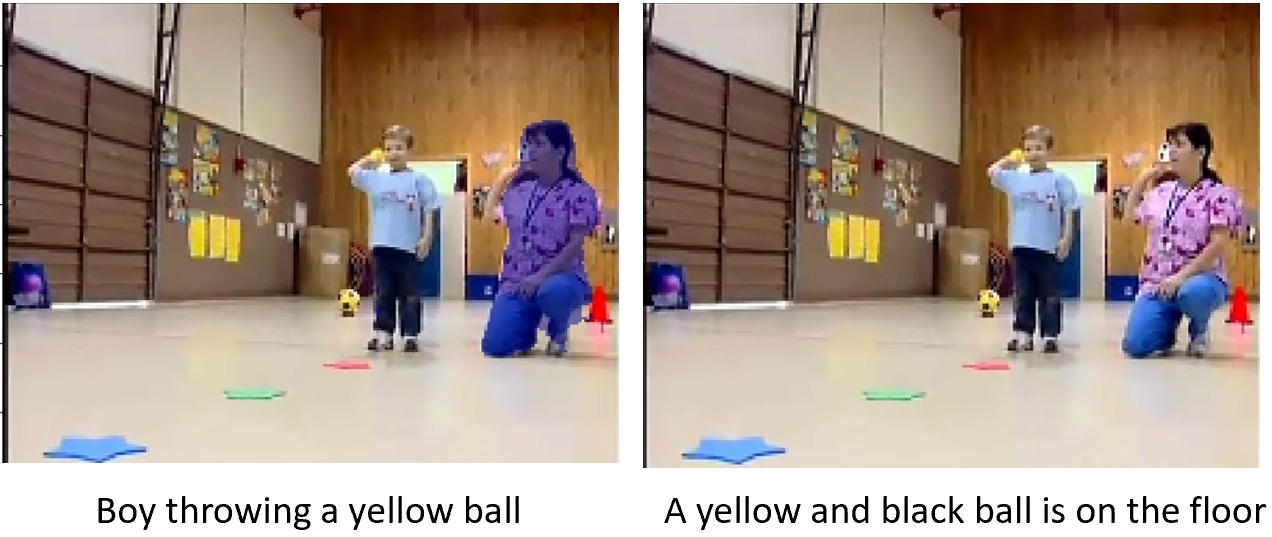}
   \caption{Failure Cases. These are frames from the same video in which different text queries resulted in distinct segmentation failure cases. In the first example the network chooses the wrong actor based on the query; in the second, the network is unable to find the queried actor.} 
\label{fig:failure}
\end{figure}

\subsection{Ablation Studies}
The ablation experiments were trained and evaluated using the pixel-level segmentations from the A2D dataset. All ablation results can be found in Table \ref{ablations-table}.
\squeezeup
\paragraph{Skip Connections} To understand the effectiveness of the parameterized skip connections from the I3D encoder, an experiment was run with these skip connections removed. This resulted in a 3\% reduction in mean IoU score and mean average precision. The decrease in performance shows that the skip connections are necessary for the network to preserve fine-grained details from the input video.

\squeezeup
\paragraph{Classification and Masking} We test the influence of the classification loss for this segmentation task, by running an experiment without back-propogating this loss. Without classification, the masking procedure would fail at test time, so masking is not used and all poses are passed forward to the upsampling network. This performed worse than the baseline in all metrics, which shows the importance of classification loss and masking when training this capsule network. To further investigate the effects of masking, we perform an experiment with no masking, but with the classification loss. Surprisingly, it performs worse than the network without masking nor classification loss; this signifies that classification loss can be detrimental to this segmentation task, if there is no masking to guide the flow of the segmentation loss gradient.


\squeezeup
\paragraph{Multi-Resolution Segmentation Loss} The base network computes a segmentation loss not only at the final output, but at multiple intermediate resolutions ($28 \times 28$, $56 \times 56$, and $112 \times 112$). This approach was used in \cite{gavrilyuk2018actor}, to great success. As an ablation, we trained the network ignoring the intermediate segmentation losses, which produced similar results to the baseline network. Thus, the multi-resolution segmentation loss has no noticeable effect on our network.


\squeezeup
\paragraph{Alternative Conditioning} We run a series of experiments to test the effectiveness of our multi-modal capsule routing procedure. We test the four other conditioning methods described in Section \ref{sect-multimodal}: the two trivial approaches (concatenation and multiplication), and the two methods which apply dynamic filtering to the video capsules (filtering the pose matrices and filtering the activations). The results suggest that multi-modal capsule routing is an effective method for merging different data modalities, since the convolution-based approaches perform much worse in all metrics. Moreover, these experiments show that it is non-trivial task to extend techniques developed for CNNs, like dynamic filtering for natural language conditioning, to capsule networks. 

\begin{table}
\small
\centering
\begin{tabular}{lccc}
\toprule
& P@0.5 & mAP & Mean IoU \\

\cmidrule(lr){2-4}
No skip connections & 49.5  & 26.9  & 43.1 \\  
\midrule
No $L_c$ nor Masking   & 49.4   & 28.8 & 43.6 \\

No Masking (with $L_c$) & 48.3  & 27.8  & 42.5 \\
\midrule
Single Resolution $L_s$ & 51.9  & 31.2  & 45.0 \\
\midrule
Concatenation & 22.9 & 9.9 & 25.0 \\
Multiplication & 38.4 & 19.4 & 35.0 \\
Filter Poses & 49.1 & 29.1 & 42.7 \\
Filter Activations & 48.8 & 29.2 & 43.0 \\
\midrule
Our Network & 52.6 & 30.3 & 46.0 \\
\bottomrule
\end{tabular}
\caption{Ablations on the A2D dataset with sentences. We test the effect of parameterized skip connections, capsule masking, the classification loss, and the multi-resolution segmentation loss on our network. We also test conventional conditioning methods on our capsule network to evaluate the effectiveness of the proposed multi-modal capsule routing procedure. The final row contains the results of our network without any changes.}
\label{ablations-table}
\end{table} 

\subsection{Failure Cases}
We find that the network has two main failure cases: (1) the network incorrectly selects an actor which is not described in the query, and (2) the network fails to segment anything in the video. Figure \ref{fig:failure} contains examples of both cases. The first case occurs when the text query refers to an actor/action pair and multiple actors are doing this action or the video is cluttered with many possible actors from which to choose. This suggests that an improved video encoder which extracts better video feature representations and creates more meaningful video capsules could reduce the number of these incorrect segmentations. The second failure case tends to occur when the queried object is small, which is often the case with the ``ball" class or when the actor of interest is far away.

\section{Conclusion and Future Work}
In this work, we propose a capsule network for localization of actor and actions based on a textual query. The proposed framework makes use of capsules for both video as well as textual representation. We introduce the concept of multi-modal capsule networks, through multi-modal EM routing for the localization of actors and actions in video, conditioned on a textual query. The existing annotations on the A2D dataset are for single frames and we extended the dataset with annotations for all the frames to validate the performance of our proposed approach. In our experiments, we demonstrate the effectiveness of multi-modal capsule routing, and observe an improvement in the performance when compared to the state-of-the art approaches. We found the capsule representation to be effective for both visual and text modalities; we plan to explore the interplay between these modalities using capsules and also apply it to other domains in future work.

\subsection*{Acknowledgement}
This research is based upon work supported by the Office of the Director of National Intelligence (ODNI), Intelligence Advanced Research Projects Activity (IARPA), via IARPA R\&D Contract No. D17PC00345. The views and conclusions contained herein are those of the authors and should not be interpreted as necessarily representing the official policies or endorsements, either expressed or implied, of the ODNI, IARPA, or the U.S. Government. The U.S. Government is authorized to reproduce and distribute reprints for Governmental purposes notwithstanding any copyright annotation thereon.

{\small
\bibliographystyle{ieee}
\bibliography{egbib}

\begin{thebibliography}{10}\itemsep=-1pt

\bibitem{carreira2017quo}
J.~Carreira and A.~Zisserman.
\newblock Quo vadis, action recognition? a new model and the kinetics dataset.
\newblock In {\em 2017 IEEE Conference on Computer Vision and Pattern
  Recognition (CVPR)}, pages 4724--4733. IEEE, 2017.

\bibitem{duarte2018videocapsulenet}
K.~Duarte, Y.~S. Rawat, and M.~Shah.
\newblock Videocapsulenet: A simplified network for action detection.
\newblock In {\em Advances in Neural Information Processing Systems}, 2018.

\bibitem{gao2017tall}
J.~Gao, C.~Sun, Z.~Yang, and R.~Nevatia.
\newblock Tall: Temporal activity localization via language query.
\newblock {\em arXiv preprint arXiv:1705.02101}, 2017.

\bibitem{gavrilyuk2018actor}
K.~Gavrilyuk, A.~Ghodrati, Z.~Li, and C.~G. Snoek.
\newblock Actor and action video segmentation from a sentence.
\newblock In {\em Proceedings of the IEEE Conference on Computer Vision and
  Pattern Recognition}, pages 5958--5966, 2018.

\bibitem{ava}
C.~Gu, C.~Sun, S.~Vijayanarasimhan, C.~Pantofaru, D.~A. Ross, G.~Toderici,
  Y.~Li, S.~Ricco, R.~Sukthankar, C.~Schmid, et~al.
\newblock Ava: A video dataset of spatio-temporally localized atomic visual
  actions.
\newblock {\em CVPR}, 2018.

\bibitem{hendricks2017localizing}
L.~A. Hendricks, O.~Wang, E.~Shechtman, J.~Sivic, T.~Darrell, and B.~Russell.
\newblock Localizing moments in video with natural language.
\newblock In {\em Proceedings of the IEEE International Conference on Computer
  Vision (ICCV)}, pages 5803--5812, 2017.

\bibitem{arsurvey}
S.~Herath, M.~Harandi, and F.~Porikli.
\newblock Going deeper into action recognition: A survey.
\newblock {\em Image and vision computing}, 60:4--21, 2017.

\bibitem{hinton2018emrouting}
G.~Hinton, S.~Sabour, and N.~Frosst.
\newblock Matrix capsules with em routing.
\newblock 2018.

\bibitem{hinton2011transforming}
G.~E. Hinton, A.~Krizhevsky, and S.~D. Wang.
\newblock Transforming auto-encoders.
\newblock In {\em International Conference on Artificial Neural Networks},
  pages 44--51. Springer, 2011.

\bibitem{tcnn}
R.~Hou, C.~Chen, and M.~Shah.
\newblock Tube convolutional neural network (t-cnn) for action detection in
  videos.
\newblock In {\em IEEE International Conference on Computer Vision}, 2017.

\bibitem{hu2016segmentation}
R.~Hu, M.~Rohrbach, and T.~Darrell.
\newblock Segmentation from natural language expressions.
\newblock In {\em European Conference on Computer Vision}, pages 108--124.
  Springer, 2016.

\bibitem{jhmdb}
H.~Jhuang, J.~Gall, S.~Zuffi, C.~Schmid, and M.~J. Black.
\newblock Towards understanding action recognition.
\newblock In {\em Computer Vision (ICCV), 2013 IEEE International Conference
  on}, pages 3192--3199. IEEE, 2013.

\bibitem{ji2018end}
J.~Ji, S.~Buch, A.~Soto, and J.~C. Niebles.
\newblock End-to-end joint semantic segmentation of actors and actions in
  video.
\newblock In {\em Proceedings of the European Conference on Computer Vision
  (ECCV)}, pages 702--717, 2018.

\bibitem{tubelets}
V.~Kalogeiton, P.~Weinzaepfel, V.~Ferrari, and C.~Schmid.
\newblock Action tubelet detector for spatio-temporal action localization.
\newblock In {\em ICCV-IEEE International Conference on Computer Vision}, 2017.

\bibitem{kay2017kinetics}
W.~Kay, J.~Carreira, K.~Simonyan, B.~Zhang, C.~Hillier, S.~Vijayanarasimhan,
  F.~Viola, T.~Green, T.~Back, P.~Natsev, et~al.
\newblock The kinetics human action video dataset.
\newblock {\em arXiv preprint arXiv:1705.06950}, 2017.

\bibitem{kazemzadeh2014referitgame}
S.~Kazemzadeh, V.~Ordonez, M.~Matten, and T.~Berg.
\newblock Referitgame: Referring to objects in photographs of natural scenes.
\newblock In {\em Proceedings of the 2014 conference on empirical methods in
  natural language processing (EMNLP)}, pages 787--798, 2014.

\bibitem{kingma2014adam}
D.~P. Kingma and J.~Ba.
\newblock Adam: A method for stochastic optimization.
\newblock {\em arXiv preprint arXiv:1412.6980}, 2014.

\bibitem{lalonde2018capsules}
R.~LaLonde and U.~Bagci.
\newblock Capsules for object segmentation.
\newblock {\em arXiv preprint arXiv:1804.04241}, 2018.

\bibitem{li2017person}
S.~Li, T.~Xiao, H.~Li, B.~Zhou, D.~Yue, and X.~Wang.
\newblock Person search with natural language description.

\bibitem{li2017tracking}
Z.~Li, R.~Tao, E.~Gavves, C.~G. Snoek, A.~W. Smeulders, et~al.
\newblock Tracking by natural language specification.
\newblock In {\em CVPR}, volume~1, page~5, 2017.

\bibitem{lin2014microsoft}
T.-Y. Lin, M.~Maire, S.~Belongie, J.~Hays, P.~Perona, D.~Ramanan,
  P.~Doll{\'a}r, and C.~L. Zitnick.
\newblock Microsoft coco: Common objects in context.
\newblock In {\em European conference on computer vision}, pages 740--755.
  Springer, 2014.

\bibitem{mikolov2013w2v}
T.~Mikolov, I.~Sutskever, K.~Chen, G.~S. Corrado, and J.~Dean.
\newblock Distributed representations of words and phrases and their
  compositionality.
\newblock In {\em Advances in neural information processing systems}, pages
  3111--3119, 2013.

\bibitem{paszke2017automatic}
A.~Paszke, S.~Gross, S.~Chintala, G.~Chanan, E.~Yang, Z.~DeVito, Z.~Lin,
  A.~Desmaison, L.~Antiga, and A.~Lerer.
\newblock Automatic differentiation in pytorch.
\newblock 2017.

\bibitem{sabour2017dynamicrouting}
S.~Sabour, N.~Frosst, and G.~E. Hinton.
\newblock Dynamic routing between capsules.
\newblock In {\em Advances in Neural Information Processing Systems}, pages
  3856--3866, 2017.

\bibitem{Shi_2018_ECCV}
H.~Shi, H.~Li, F.~Meng, and Q.~Wu.
\newblock Key-word-aware network for referring expression image segmentation.
\newblock In {\em The European Conference on Computer Vision (ECCV)}, September
  2018.

\bibitem{sigurdsson2016hollywood}
G.~A. Sigurdsson, G.~Varol, X.~Wang, A.~Farhadi, I.~Laptev, and A.~Gupta.
\newblock Hollywood in homes: Crowdsourcing data collection for activity
  understanding.
\newblock In {\em European Conference on Computer Vision}, pages 510--526.
  Springer, 2016.

\bibitem{twostream}
K.~Simonyan and A.~Zisserman.
\newblock Two-stream convolutional networks for action recognition in videos.
\newblock In {\em Advances in neural information processing systems}, pages
  568--576, 2014.

\bibitem{ucf101}
K.~Soomro, A.~R. Zamir, and M.~Shah.
\newblock Ucf101: A dataset of 101 human actions classes from videos in the
  wild.
\newblock {\em arXiv preprint arXiv:1212.0402}, 2012.

\bibitem{xu2015can}
C.~Xu, S.-H. Hsieh, C.~Xiong, and J.~J. Corso.
\newblock Can humans fly? action understanding with multiple classes of actors.
\newblock In {\em Proceedings of the IEEE Conference on Computer Vision and
  Pattern Recognition}, pages 2264--2273, 2015.

\bibitem{yamaguchi2017spatio}
M.~Yamaguchi, K.~Saito, Y.~Ushiku, and T.~Harada.
\newblock Spatio-temporal person retrieval via natural language queries.
\newblock {\em arXiv preprint arXiv:1704.07945}, 2017.

\bibitem{zhao2018investigating}
W.~Zhao, J.~Ye, M.~Yang, Z.~Lei, S.~Zhang, and Z.~Zhao.
\newblock Investigating capsule networks with dynamic routing for text
  classification.
\newblock {\em arXiv preprint arXiv:1804.00538}, 2018.

\end{thebibliography}
}

\newpage

\appendix
\section{Appendices}

Here we include many qualitative results, and quantitative results which could not be included in the main text. Also, we include figures and a more in-depth description of the network architecture.


\section{Network Architecture}
\begin{figure}[t]
\begin{center}
\includegraphics[width=0.95\linewidth]{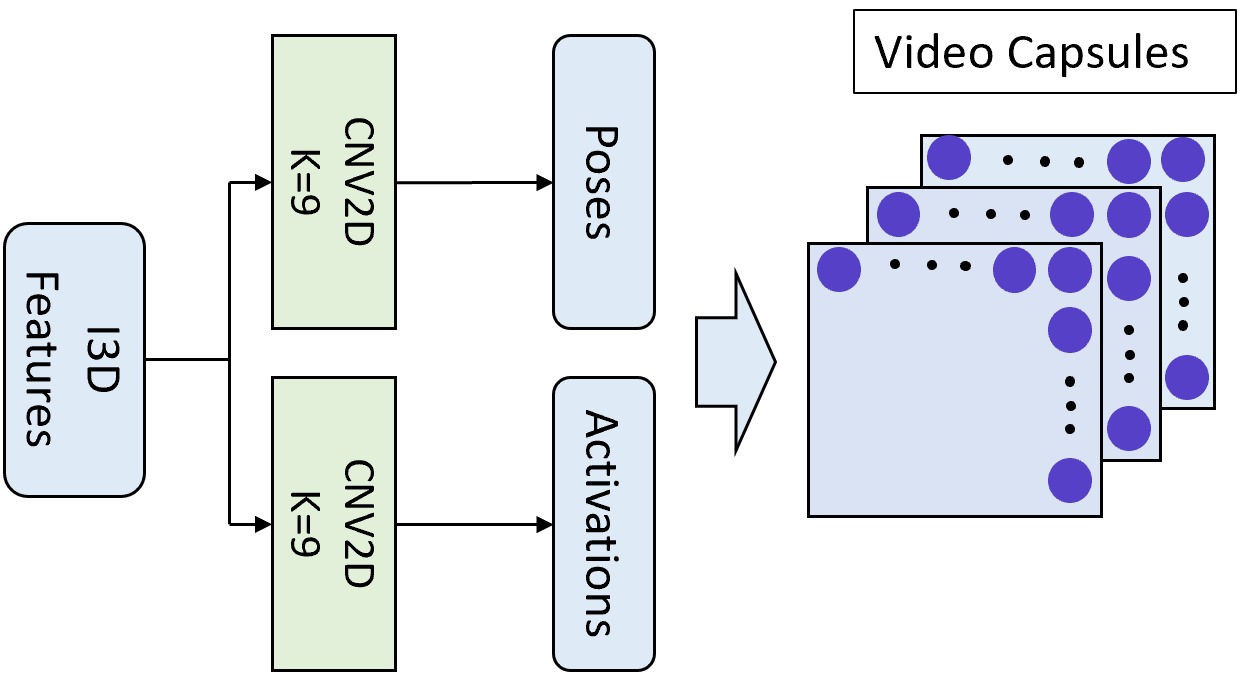}
\end{center}
   \caption{Video Capsule Network. Video capsules are formed from the I3D features by convolution with one layer to create the 4x4 pose matrix for each capsule, and another layer to create the activation for each capsule.}
\label{fig:videocapsule}
\end{figure}

\begin{figure}[t!]
\begin{center}
\includegraphics[width=0.95\linewidth]{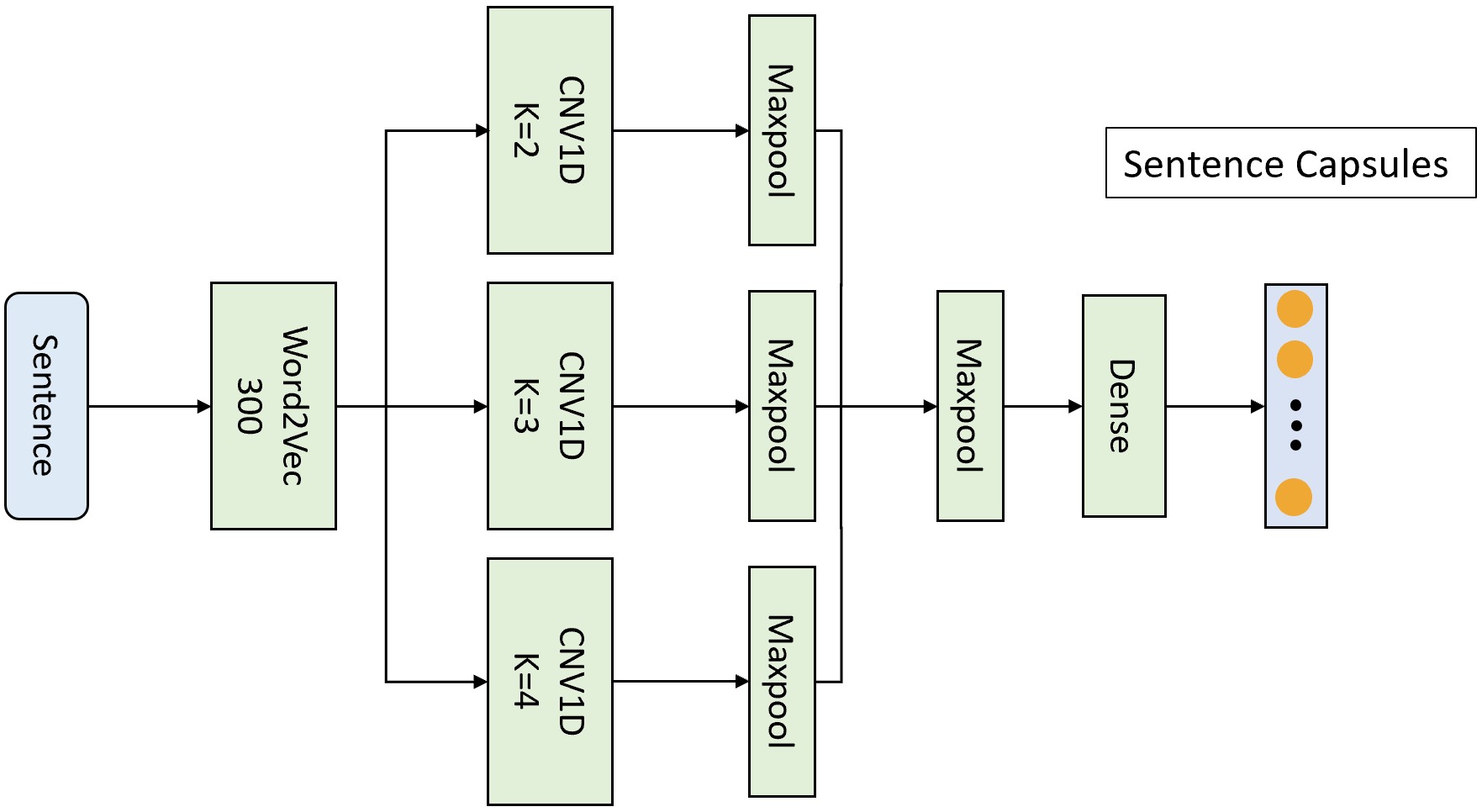}
\end{center}
\caption{Sentence Network. Each word from a natural language sentence is converted into a size 300 word2vec vector. The vectors go through a convolutional network and are then reshaped into the poses and activations of capsules representing the sentence.  }
\label{fig:sentencecapsule}
\end{figure}

\begin{figure}[t]
\begin{center}
\includegraphics[width=0.95\linewidth]{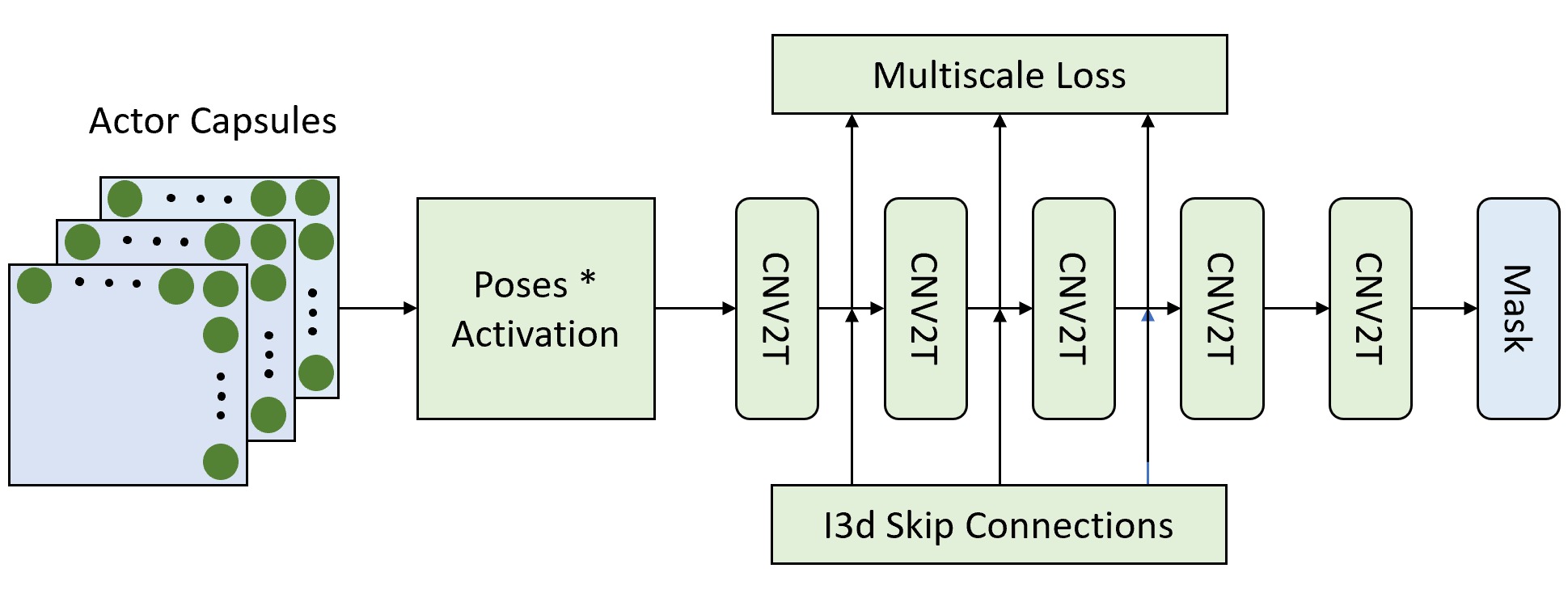}
\end{center}
   \caption{Upsampling Network. The pose matrices undergo a masking procedure and are passed through a series of convolutional transpose layers to create the binary segmentation mask. Skip connection from the I3D are made at 3 different resolutions. Segmentation loss is backpropagated from the final output as well as 3 intermediate levels.}
\label{fig:upsampling}
\end{figure}

When designing our network, we found several components key to obtaining our state-of-the-art results. We  explain some of these components, and how they impacted our results. Furthermore, we include several figures (\ref{fig:videocapsule}, \ref{fig:sentencecapsule}, \ref{fig:upsampling}) which illustrate the construction of the video capsules, the sentence network, and the upsampling network respectively.

\subsection{Video Encoder}
Our network uses a pretrained I3D network to encode the video into a set of feature maps of size $28 \times 28$. Originally we used the simpler C3D network \cite{tran2015learning}, to encode the feature maps, but it achieved substantially worse results - a mean IoU of 34\%. This shows that the features extracted from the I3D network, are more useful for our video capsules than were the features from the C3D network. Not only that, but it suggests that future improvements in video feature extraction techniques and networks, will lead to improved capsule representations and improved results.

\subsection{Sentence Network}
We also tested several different configurations for our sentence network. We began with using the capsule networks (both Capsule-A and Capsule-B) described in Figure 2 of \cite{zhao2018investigating} which showed strong results on text classification. These network have several capsule layers, but their use led to poor results: a mean IoU of 35.7\% for the Capsule-A network, and a mean IoU of 36.4\% for the Capsule-B network. Although these network performed well on the text classification task, a different set of text features must be learned to condition visual features. Therefore, our sentence network with conventional convolutional layers, max-pooling, and a fully connected layer, is better able to extract textual features for the task of video segmentation from a sentence.

\section{Evaluations}
We include several tables which we were unable to include in our main text. Table \ref{bbox-table} shows the results of our network trained on the bounding box annotations from A2D, and tested on JHMDB. The network's outputs are more box-like because it was trained on bounding boxes as opposed to pixel-wise segmentations; therefore, the network has strongest results when tested using bounding box ground-truths. Tables \ref{ablations-table} and \ref{referit-table} contain all the metrics which we were unable to include for the ablation experiments and ReferItGame experiments, respectively.

\section{Qualitative Results}
We have generated several videos with the segmentations produced by our networks for both the A2D and JHMDB datasets. Each video contains the input sentences, color coded to match the ground-truth colors. The first row of each video has the ground-truth bounding boxes (for A2D) or ground-truth segmentation masks (for JHMDB). The second row is the output of the network which was trained on the key frames of the A2D dataset, which had pixel-wise segmentation ground-truths. The third row is the output of the network which was trained on all the frames of the A2D dataset, using bounding-box ground-truths. In our analysis of the qualitative results, we will refer to the prior network as the "Key Frame network" and the latter network as the "Bounding Box network".

\subsection{Single Actor}
The networks seem to perform best when there is a single actor in the scene. If this is the case, we find that the Key Frame network produces very fine-grained segmentations which maintain the boundaries of the actors; Meanwhile, the Bounding Box network successfully segment a box around the actor. This behaviour can be observed in the videos in the A2DSingleActor folder, which contains examples from the A2D dataset, and the JHMDB folder, which contains examples from the JHMDB dataset.

\subsection{Multiple Actors}
The network can also perform well with multiple actors in the scene as seen in the A2DMultiActor videos. In these cases, the segmentations are not as precise, but the general location of the actors is being segmented by both networks. We note that the Bounding Box network, which was trained with all the frames of dataset, tends to produce more consistant multi-actor segmentations: as seen in videos "video3\_multi\_a2d" and "video5\_multi\_a2d". In both of these cases, each instance is of the same actor class, and the Key Frame network seems to incorrect segment one of the instances.

\subsection{Failure Cases}
As mentioned in the main text, our network tends to fail on the A2D dataset when there are multiple instances of the same actor class. We present 5 videos in which our network fails in the A2DFailure folder. The probability of failure is increased when the sentence queries are vague, or could describe multiple actors within the scene, like in the video "video1\_failure\_a2d". Several birds can fit the description of "sparrow sitting on the grass" or "sparrow is walking on the brown grass". In these cases it would be very difficult for even a human to correctly segment the video. In many cases, the failure occurs when there are many similar actors near each other, like in the videos "video2\_failure\_a2d.mp4" and "video5\_failure\_a2d". In the first, there are multiple people running next to each other, while the second contains several cars moving near each other.

Since the JHMDB dataset only has a single actor in each video, the failures encountered are not from "difficult" queries or videos, but rather a result of the mismatch between the training and testing data. A2D videos tend to have humans perform an action requiring large amounts of motion - like walking, running, jumping or rolling - while the A2D videos have many videos in which the action has little motion - like brushing hair or archery. Thus, both the videos, and the input textual queries, are quite different between the datasets, which can cause a large performance discrepancy during evaluation. Examples of failure cases on the JHMDB dataset can be found in the JHMDBFailure folder.

\begin{table*}
\begin{center}
\begin{tabular}{l ccccc c cc} 
\toprule
 & \multicolumn{5}{c}{\bf Video Overlap} & \multicolumn{1}{c}{\bf v-mAP} & \multicolumn{2}{c}{\bf Video IoU} \\
 & P@0.5 & P@0.6 & P@0.7 & P@0.8 & P@0.9 & 0.5:0.95 & Overall & Mean \\
\cmidrule(lr){2-6} \cmidrule(lr){7-7} \cmidrule(lr){8-9} 
All frames         & 16.3 &  2.0 & 0.1 & 0.0  & 0.0  & 2.3  & 37.9  & 35.6 \\
All frames  (bbox) & 46.7 &  32.1 & 15.8 & 3.2  & 0.0  & 16.3  & 44.5  & 44.4 \\  
\bottomrule
\end{tabular}
\end{center}
\caption{Results for network trained on the bounding box annotations for A2D with sentences, and evaluated on JHMDB. The first row is the network tested against the ground-truth pixel-wise segmentations from the JHMDB dataset. The second row is the network tested against bounding boxes around the ground-truth segmentations from the JHMDB dataset. Since our network was trained with bounding boxes, it performs better when it is evaluated against bounding box ground-truths.}
\label{bbox-table}
\end{table*}

\begin{table*}
\begin{center}
\begin{tabular}{l ccccc c cc}
\toprule
 & \multicolumn{5}{c}{\bf Overlap} & \multicolumn{1}{c}{\bf mAP} & \multicolumn{2}{c}{\bf IoU} \\
 & P@0.5 & P@0.6 & P@0.7 & P@0.8 & P@0.9 & 0.5:0.95 & Overall & Mean \\
\cmidrule(lr){2-6} \cmidrule(lr){7-7} \cmidrule(lr){8-9} 
No skip connections & 49.5  & 41.2  & 29.1  & 14.6  & 1.5  & 26.9  & 56.7  & 43.1 \\  
\midrule
No $L_c$ nor Masking   & 49.4  & 42.5  & 32.7  & 19.6  & 3.3  & 28.8  & 57.6  & 43.6 \\

No Masking (with $L_c$) & 48.3  & 41.4  & 31.2  & 18.4  & 3.1  & 27.8  & 56.6  & 42.5 \\
\midrule
Single Resolution $L_s$ & 51.9  &  45.2  & 35.4  & 21.7  & 4.2  & 31.2  & 59.9  & 45.0 \\
\midrule
Concatenation & 22.9 & 15.4 & 7.6 & 2.0 & 0.1 & 9.9 & 35.1 & 25.0 \\
  Multiplication & 38.4 & 30.1 & 20.9 & 9.7 & 0.8 & 19.4 & 48.2 & 35.0 \\
  Filter Poses & 49.1 & 42.3 & 32.6 & 19.1 & 3.2 & 29.1 & 57.2 & 42.7 \\
  Filter Activations & 48.8 & 42.7 & 33.4 & 20.1 & 3.8 & 29.2 & 56.8 & 43.0 \\
\midrule
{\bf Our Network}  & {\bf 52.6}  & {\bf 45.0}  & {\bf 34.5}  & {\bf 20.7}  & {\bf 3.6}  & {\bf 30.3}  & {\bf 56.8}  & {\bf 46.0} \\
\bottomrule
\end{tabular}
\end{center}
\caption{All metrics for the ablations trained and tested on the A2D dataset with sentences. We test the effect of parameterized skip connections, capsule masking, the classification loss, and the multi-resolution segmentation loss on our network. We also test conventional conditioning methods on our capsule network to evaluate the effectiveness of the proposed multi-modal capsule routing procedure. The final row contains the results of our network without any changes.}
\label{ablations-table}
\end{table*}

\begin{table*}
\begin{center}
\begin{tabular}{l ccccc c}
\toprule
 & Overall IoU & P@0.5 & P@0.6 & P@0.7 & P@0.8 & P@0.9   \\
\cmidrule(lr){2-6} \cmidrule(lr){7-7}
Hu et al.* \cite{hu2016segmentation} & 56.83 & 43.86 & 35.75 & 26.65 & 16.75 & 6.47  \\
Shi et al. \cite{Shi_2018_ECCV} & 59.09 & 45.87 & 39.80 & 32.82 & 23.81 & 11.79  \\
\midrule
Our Network         & 55.7  & 43.4  & 36.2  & 28.3  & 19.6  & 9.7   \\   
\bottomrule
\end{tabular}
\end{center}
\caption{Results on ReferItGame dataset. *This result is using Deeplab101 as a backbone, as described in \cite{Shi_2018_ECCV}. We achieve comparable results, even with a network designed for video inputs.}
\label{referit-table}
\end{table*}

\end{document}